\documentclass[12pt]{article}
\usepackage{arxiv}

\usepackage{cite}
\usepackage{amsmath,amssymb,amsfonts}
\usepackage{algorithmic}
\usepackage{graphicx}
\usepackage{textcomp}
\usepackage{xcolor}
\usepackage{amsmath,amssymb}
\usepackage{algorithmic}
\usepackage{graphicx}
\usepackage{textcomp}
\usepackage{xcolor}
\usepackage{amsfonts}

\usepackage[colorinlistoftodos]{todonotes}
\usepackage{hyperref}
\usepackage{float}
\usepackage{multirow}
\usepackage{booktabs}
\usepackage{grffile}
\usepackage{color}
\usepackage{bm}
\usepackage{float}
\usepackage{xspace}
\usepackage{multirow}
\usepackage{hyperref}
\usepackage{mwe}
\usepackage{subfig}
\usepackage[ruled,vlined]{algorithm2e}

\def\BibTeX{{\rm B\kern-.05em{\sc i\kern-.025em b}\kern-.08em
    T\kern-.1667em\lower.7ex\hbox{E}\kern-.125emX}}
\date{}

\begin{document}
\title{Deep Learning with Kernel Flow Regularization for Time Series Forecasting
}


\author{ {Mahdy~Shirdel} \\
	Machine Learning Manager\\
	Beyond Limits Inc \\
	\texttt{mshirdel@beyond.ai} \\
	\And
	{Reza Asadi} \\
	Data Scientist\\
	Beyond Limits Inc \\
	\texttt{rasadi@beyond.ai} \\
	\And 
	{Duc Do} \\
	Machine Learning Engineer\\
	Beyond Limits Inc\\
	\texttt{ddo@beyond.ai} \\
	\And
	{Micheal Hintlian} \\
	Data Scientist Intern with Beyond Limits Inc\\
	PhD student with University of Southern California \\
	\texttt{michael.hintlian@gmail.com} \\
}

\maketitle

\begin{abstract}
Long Short-Term Memory (LSTM) neural networks have been widely used for time series forecasting problems. However, LSTMs are prone to overfitting and performance reduction during test phases. Several different regularization techniques have been shown in literature to prevent overfitting problems in neural networks. In this paper, first, we introduce application of kernel flow methods for time series forecasting in general. Afterward, we examine the effectiveness of applying kernel flow regularization on LSTM layers to avoid overfitting problems. We describe a regularization method by applying kernel flow loss function on LSTM layers. In experimental results, we show that kernel flow outperforms baseline models on time series forecasting benchmarks. We also compare the effect of dropout and kernel flow regularization techniques on LSTMs. The experimental results illustrate that kernel flow achieves similar regularization effect to dropout. It also shows that the best results is obtained using both kernel flow and dropout regularizations with early stopping on LSTM layers on some time series datasets (e.g. power-load demand forecasts).
\end{abstract}



\section{Introduction} 

Time series data and time series forecasting are essential parts of many different industrial applications \cite{tealab2018time} for decision making or operation safety, such as in power load supply and demand, energy systems, finances and etc. In a time series forecasting problem, the model forecasts a window of future time stamps based on the observed historical data. Following articles are some of the aforementioned applications in power load forecasting \cite{din2017short}, traffic forecasting \cite{tian2018lstm}, exchange-rates forecasting \cite{bou2020forecasting} and so on. 

Traditional forecasting approaches such as exponential smoothing and autoregressive models have been widely studied and applied. Besides being highly interpretable, they can achieve high forecasting accuracy especially when the amount of observations is small. However, in many real world applications nowadays, time series data come in high volumes from multiply sources, thus they tend to have both short-term and long-term temporal patterns embedded in non-linear relations among multiply variables. It is challenging for the classical approaches to model and leverage such complex temporal patterns. Recurrent Neural Networks (RNNs), along with its popular variant LSTM, have been used to address these challenges \cite{benidis2020neural}.

LSTM has been very successful in NLP tasks such as machine translation \cite{yang2020survey}, and it also shows promising results in time-series forecasting problems \cite{du2020multivariate}. Different RNN variants were compared in \cite{bianchi2017overview} and LSTM is shown to have best prediction accuracy for highly non-linear sequences with sharp spikes and dips due to its memory cell modification mechanism. An LSTM model proposed in \cite{salinas2020deepar} where LSTM's hidden states were leveraged to predict parameters of Bayesian models for future time-steps. The model was shown to drastically improve forecast accuracy with minimum or no hyperparameter tuning on a wide variety of datasets. In another study \cite{lai2018modeling}, LSTM blocks were combined with 1D convolutional layer to capture both short-term and long-term patterns in the non-linear data. 1-dimensional convolutional layers were also used to encode short-term local information into low dimensional vectors, then LSTM was used to decode the vector. The approach is shown to improve state-of-the-arts results in multiple benchmark time series forecasting datasets. 

One of the drawbacks of working with deep learning models, including LSTM, is to have an overfitted model \cite{srivastava2014dropout}. Deep learning models are designed to have millions of trainable parameters and learn very complex patterns. However, these models get overfitted in small datasets. Unlike computer vision and NLP, time series data are usually smaller, with less complexities. While deep learning models are shown effective for time series forecasting problems, they can easily be overfitted, and so regularization methods are necessary to prevent overfitting. 

The regularization methods, such as L1- and L2-norm, can be applied to each layer to get sparse features, and prevent overfitting \cite{xu2019overfitting}. Dropout, introduced in \cite{srivastava2014dropout}, is shown to be effective in wide range of deep learning models. Spectral dropout, introduced in \cite{khan2019regularization}, eliminate neural network activations based on weak and noisy Fourier domain coefficients. 

Kernel flow method is studied and proposed in \cite{owhadi2019kernel}, where the method approximates a function by interpolating training data. In \cite{yoo2021deep}, they study kernel flow regularization on neural networks, and show that the model with kernel flow regularization method outperforms the model with dropout regularization. They apply kernel flow regularization on convolutional neural networks and images. Here, we apply kernel flow regularization on LSTM layers and for time series forecasting problems. We show that with properly designed network and hyper parameter tuning, kernel flow regularization has a comparable regularization effect on neural networks with dropout. Additionally, the benefit of using kernel flow regularizer is the deterministic nature of this method for the regularization and better generalization for future unseen datasets. We also show that in time series forecasting problems on four benchmark datasets, the combination of kernel flow and dropout regularizers achieve better performance. The summary of our contribution is as follows:

\begin{itemize}

    \item We illustrate kernel flow method for time series forecasting  can achieve comparable results to baseline models in a wide range of benchmarked datasets.
    
    \item We analyze the effect of kernel flow regularization on LSTM layers and for time series forecasting problem.

    \item On three time series benchmark datasets, we show that kernel flow regularization prevents the model from overfitting by evaluating the training and testing error during the training process.

    \item We show that an LSTM model with both kernel flow regularization and dropout achieves an improvement compared to an LSTM model with only dropout regularization.

\end{itemize}

\section{Technical Backgrounds}

\subsection{Multi-variate time series forecasting problem}

A multi-variate time series data $\bm{X}$, is represented with a 2-dimensional matrix of size $(\bm{n}, \bm{d})$, where $\bm{n}$ is the number of time stamps and $\bm{d}$ is the number of features. For a time series forecasting problem, we use a sliding-window method, which generates  data points for each time stamp. A data point includes the last, look-back, $\bm{l}$ time stamps as the input, and the next, horizon, $\bm{h}$ time stamps as the output. The input data point $\bm{x}^{\bm{t}}$ is a 2-dimensional matrix of size $(\bm{l}, \bm{d})$, and the target data point, $\bm{y}^{\bm{t}}$, is a 2-dimensional matrix of size $(\bm{h}, \bm{d})$, obtained for time stamp $\bm{t}$. If the goal of the problem is to forecast some of the input features, then, without any changes in the problem formulation, $\bm{y}^{\bm{t}}$ has a size of $(\bm{h}, \bm{s})$, where $\bm{s}$ is the number of subset features.

A machine learning model receives the input $\bm{x}^{\bm{t}}$ and forecast the target $\bm{y}^{\bm{t}}$ for all $\bm{t} \in [1, \dots, \bm{n}]$. The values of time series data $\bm{x}$ are numerical, and can be scaled in the range of $[-1,+1]$. This is a multi-input/multi-output regression problem.

\subsection{Kernel flow}
Kernel flow is an algorithm for selecting family of kernels for Kriging/Gasussian processing regression, introduced in \cite{owhadi2019kernel}. The method learn the kernel to approximate an unknown function $\bm{u}(.): \mathbb{R} \rightarrow \mathbb{R}$, which receives a dataset with an input $\bm{x}_{i}$ and output $\bm{y}_{i}$. A family of non-degenerate kernels $\bm{K}(\bm{x}, \bar{\bm{x}})$ approximates the mapping function $\bm{u}(.)$ by the following interpolant:

\begin{equation}
    \bm{u}(\bm{x}) = \bm{K}(\bm{x}, \bm{X})\bm{K}(\bm{X}, \bm{X})^{-1}\bm{Y}
\end{equation}

where $\bm{X}$ is the list of input data points, and $\bm{Y}$ is the list of output data points. 

Assume a batch of data points is represented with $(\bm{X}^{\bm{b}}, \bm{Y}^{\bm{b}})$. The half of the batch is randomly selected and represented with $(\bm{X}^{\bm{c}}, \bm{Y}^{\bm{c}})$. A family of kernels $\bm{K}_{\theta}(\bm{x}, \bar{\bm{x}})$ is parametrized by $\theta$. The relative error of interpolation between $\bm{X^b}$ and $\bm{X^c}$ is then computed in Eq 2, which is minizmied to find the optimal solution. 
\begin{equation}
 \rho(\bm{\theta},
             \bm{X^b},
             \bm{Y^b},
             \bm{X^c},
             \bm{Y^b}) = 
             1-\frac
             {\bm{Y^{c,T}}
             \bm{K_\theta(X^c,X^c)^{-1}} \bm{Y_c}}
             {\bm{Y^{b,T}} \bm{K_\theta(X^b,X^b)^{-1}} \bm{Y^b}}
 \end{equation} 
 The kernel flow algorithm finds an optimum value for $\theta$, which minimizes the loss function represented in Eq 2. A mini-batch gradient descent method iteratively selects the mini-batch $(\bm{X}^{\bm{b}}, \bm{Y}^{\bm{b}})$, and randomly selects the half of the batch $(\bm{X}^{\bm{c}}, \bm{Y}^{\bm{c}})$ to train the model.
 
 In summary, the method uses the loss of accuracy incurred by removing half of the data as a loss of kernel selection. The l2-norm variant of kernel flow was also proposed in \cite{yoo2021deep}. The loss function of l2-norm variant kernel flow is shown with:

\begin{equation}
    || \bm{Y}^{\bm{c}} - \bm{K}_{\theta}(\bm{X^c}, \bm{X^b})\bm{K}_{\theta}(\bm{X^b}, \bm{X^b})^{-1}\bm{Y^b}||
\end{equation}


\begin{figure*}[!ht]

\centering
    {\includegraphics[width=0.40\linewidth]{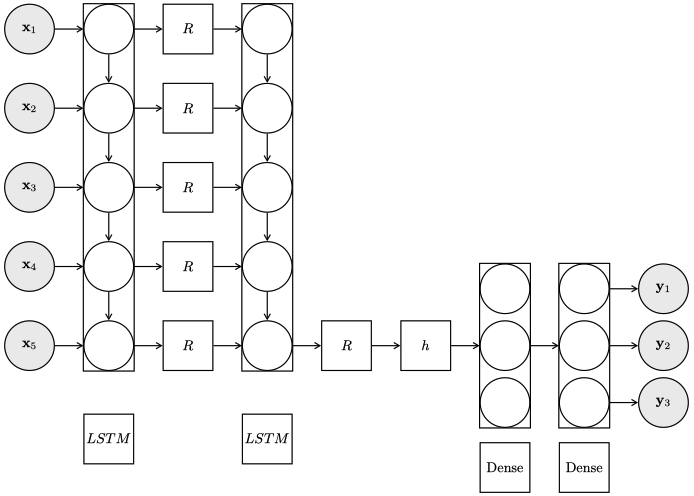}}
\caption{Stacked LSTM with Regularization methods.}\label{fig:stacked-lstm}
\end{figure*}

\section{Kernel Flow Regularization for Deep Learning}

We consider stacked LSTM for time series forecasting problems, represented in Fig. \ref{fig:stacked-lstm}. In this figure, we represent a stacked LSTM neural network, with two LSTM layers and two fully-connected layers. The boxes represented with $\bm{R}$ are the regularization functions, and we use both dropout and kernel flow regularization. The dropout function randomly eliminates a part of units in each layer. In LSTM layers, the dropout eliminates the output of lstm cells per time stamps. The kernel flow regularization applies a loss function on the units and the model optimizes kernel parameters along with the neural network parameters during training. 

Stacked LSTM is a common deep learning component, that usually can be used with other architectures such as Sequence-2-Sequence \cite{du2018time} or Attentions models \cite{du2020multivariate}. In the architecture of Stacked LSTM, each LSTM layer receives a sequence of input data, and outputs a hidden vector. The hidden vectors can be passed to Activation, Dropout and/or Batch-normalization functions, before passing to the next LSTM layer. The final LSTM layer returns the hidden vector, which is the vector representation of the input data. In other words, stacked LSTM layers convert the input time series data into its vector representation. The hidden vector is connected to multiple fully-connected layers, where each fully-connected layer is connected to the next ones by applying Activation, Dropout and/or Batch-normalization layers. The final fully-connected layer has the unit size equal to the target size, which is horizon size $\bm{h}$, multiplied by the number of output features. 

In our current model, we assume Mean-Square-Error as the loss function, which minimizes the distance between target and the predicted vector as follows: 

\begin{equation} 
\frac{1}{\bm{m}\bm{n}} \sum_{\bm{t=1}}^{\bm{d}} (\bm{y}^{\bm{t}} - \bar{\bm{y}}^{\bm{t}})^{2} 
\end{equation} 

where $\bm{y}^{\bm{t}} $ is the target value and $\bar{\bm{y}}^{\bm{t}}$ is the predicted value at time $\bm{t}$, and $\bm{m}$ is the size of output variable and $\bm{n}$ is the number of rows in time series data. To represent the loss function with batch data points, we define $\bm{b}$ as the batch size and $\bm{d}$ as the number of batches, obtained by $\frac{\bm{n}}{\bm{b}}$.

Throughout the experiments, we found out LSTM layers have the most important impact on overfitting problem. If there are dropout functions on each LSTM layer, it can prevent fitting more than other layers, such as fully-connected and convolutional layers. Hence, we add a dropout function after each LSTM layer. The dropout function removes dropout-rate p\% of the hidden vectors for each time stamp. The dropout significantly prevents overfitting in the experimental results. Moreover, our analysis shows that batch-normalization and Activation functions between LSTM layers does not improve the results.

Next, we describe kernel flow regularization on LSTM layers, represented in Algorithm  \ref{algorithm:1}. A kernel flow regularization function, $\mathsf{KernelFlowReg}(.,.)$, receives the output of an LSTM layer, represented with $\bm{x}$, with a shape of $(\bm{b}, \bm{h})$, which $\bm{b} $ is the batch-size, and $\bm{h} $ is the size of hidden vector. The kernel flow regularization layer splits the data into two parts. $\bm{x}_{f}$ is the full batch of data points, and $\bm{x}_{c}$ is the random selection of half of the data points. The parameterized kernel, $\bm{K}(.,.)$ is a family of kernels with trainable parameters $\gamma$, that receives both sets of the data points $\bm{x}_{f}$ and $\bm{x}_{c}$. The goal is to obtain the following value as it is presented in Section II.B:

\begin{equation} 
\bm{z} = \bm{K}(\bm{x}_f, \bm{x}_c) \bm{K}(\bm{x}_c, \bm{x}_c)^{-1}  
\end{equation} 

The inverse of the second kernel is obtained in $\bm{x}^{r}_{2}$. The output of layer is $\bm{y}$, returned from the neural network. If kernel flow regularization is applied to $\bm{l}$ neural network layers, then we have $\{ 1, \dots, \bm{l}\}$ output vectors.
 
 \begin{algorithm}
\SetAlgoLined 
\KwResult{output of kernel flow layer for a batch of data points $\bm{x}$} 
$\mathsf{kernelFlowReg}$ ($\bm{x}$ , $\bm{b}$):\\
$~~~~~~\bm{x}_{f} = \bm{x}[0: \bm{b}]$ \\
$~~~~~~\bm{x}_{c} = \bm{x}[0: \frac{\bm{b}}{2}]$\\ 
$~~~~~~\bm{x}_{1} = \bm{K}(\bm{x}_{f}, \bm{x}_{c})$ \\ 
$~~~~~~\bm{x}_{2} = \bm{K}(\bm{x}_{c}, \bm{x}_{c})$ \\
$~~~~~~\bm{x}^{r}_{2} = \mathsf{Inverse}(\bm{x}_{2})$ \\
$~~~~~~\bm{z} = \bm{x}_{1} * \bm{x}^{r}_{2}$ \\
~~~~~~return $\bm{z} $
\caption{kernel flow layer}\label{algorithm:1}
\end{algorithm}

During each epoch, we shuffle the data. So the first half of the data is a random set of data points. The kernel flow regularization of each layer is added to the MSE loss as follows. The following loss function: 

\begin{equation} 
	\frac{1}{\bm{m}\bm{n}} \sum_{\bm{t=1}}^{\bm{d}} (\bm{y}^{\bm{t}} - \bar{\bm{y}}^{\bm{t}})^{2} + \sum_{\bm{i}=1}^{\bm{q}} \lambda_{\bm{i}}\frac{1}{\bm{m}\bm{n}} \sum_{\bm{t}=1}^{\bm{d}} (\bm{y}^{\bm{t}} -  \bm{z}_{i}^{t}  \bm{y}_{c}^{t} )^{2}
\end{equation}

where $\lambda_{\bm{i}}$ is the weight of kernel flow loss functions (as hyper-parameters) for layer $\bm{i}$ and $\bm{q}$ is the number of neural network layers with kernel flow regularization layer. Moreover, $\bm{y}_{c}$ is the target of the random selection of half of data points.

In Algorithm \ref{algorithm:2}, we show the pseudo code of a stacked LSTM neural network with kernel flow regularization. We assume that the neural network has one LSTM layer with kernel flow regularization, and one fully-connected layer. The function $\mathsf{stackedLSTM}(.,.)$ is the neural network model which receives the input data. The output of LSTM layer is a sequential hidden vectors, passed to the kernel flow regularization layer. The last sequential hidden vector $\bm{l}_{1}^{t}$ is passed to the fully-connected layer. The output of model are two vectors, $\bm{y}$ is the neural network output, $\bm{y}_{1}$ is the output of kernel flow regularization. If there is $\bm{q}$ number of layers with kernel flow regularization, then Algorithm \ref{algorithm:2} returns $\{\bm{y}_{1} , \dots, \bm{y}_{q}\}$. The loss function, represented in Eq. 5, receives these two vectors, and using back-propagation, the mini-batch gradient descent algorithm trains the model.

\begin{algorithm}
\SetAlgoLined 
\KwResult{Neural Network's outputs} 
$\mathsf{stackedLSTM} (\bm{x} , b):$ \\
$~~~~~~\bm{l}_{1} = \mathsf{LSTM}(\bm{x})$ \\
$~~~~~~\bm{y}_{1} = \mathsf{kernel flowReg}(\bm{l}_{1}, b)$ \\
$~~~~~~\bm{l}_{1} = \bm{l}_{1}^{\bm{t}}$ \\
$~~~~~~\bm{l}_{2} = \mathsf{Dense}(\bm{l}_{1})$ \\
$~~~~~~\bm{l}_{2} = \mathsf{Relu}(\bm{l}_{2})$ \\
$~~~~~~\bm{y} = \mathsf{Dense}(\bm{l}_{2})$ \\
$return ~\bm{y}, \bm{y}_{1}$ \\

\caption{Stacked-LSTM with kernel flow regularization}
\label{algorithm:2}
\end{algorithm}

\section{Experimental Results}

In this section, we first evaluate kernel flow itself for time series forecasting problem. Then, we evaluate the performance of the proposed regularization method for deep learning models on time series benchmarks. Throughout the analysis, the best models are selected based on validation errors using early stopping method. All of the presented results in tables are testing errors.

\subsection{Datasets}

We consider the following five time series datasets to evaluate the effect of regularization on time series data.

\begin{itemize} 


\item PJM power load: this data represents hourly power load data of 28 given regional zones in middle and north-east of USA for three years \cite{PJM_data}. We added cyclical time features, such as hour, weekday, weekends, months of the year. There are 35 input time series, and 28 target time series, power-load for each regional zone.  

\item Solar data: this data is obtained from \cite{lai2018modeling}. The data represents 10-min 137 PV plants in Alabama state in 2006.

\item Electricity data: this data is obtained from \cite{lai2018modeling}. The data represents 15-min kWh from 2012 to 2014 for 321 clients.

\item Traffic data: this data is obtained from \cite{lai2018modeling}. The data represents hourly traffic occupancy of Bay Area, California for 48 months.

\item Exchange-rate data: this pre-processed data is obtained from \cite{lai2018modeling}. The daily exchange rates of eight countries from 1990 to 2006. We consider 60\% of data as the training data, and 40\% as the testing data.

 \end{itemize}

\subsection{Preprocessing} 

All the given datasets are scaled into the range of [0,1]. We consider a sliding window method with look-back variable, which considers the last few time stamps as the input, and a forecasting horizon, which considers the forecasting problem for the next few time stamps as the target. These input and output values are obtained for all the time stamps $\bm{t}$, presented in Section 2.1. We also de-trend the data by using a differencing method, by subtracting the input and output values at time $\bm{t}$. Also, we add cyclical time features for hours of the day, weekday, and month of the year.

\subsection{Evaluation metrics}

In our evaluation, we use some known metrics \cite{du2020multivariate}, such as Root-Mean-Square-Error(RMSE), Mean-Absolute-Error(MAE), Mean-Absolute-Percentage-Error(MAPE), in addition to, Root-Relative-Squared-Error(RSE) and Empirical-Correlation-Coefficient(CORR). For simplicity, we present RSE and CORR on time series data as follows: 

\begin{itemize}
	\item Root Relative Squared Error (RSE):
        \begin{equation}
        \begin{aligned}
        RSE = \frac{\sqrt{\sum_{\bm{i}} \sum_{\bm{t}} (\bm{y}_{\bm{i}}^{\bm{t}} - \bar{\bm{y}}_{\bm{i}}^{\bm{t}})^{2}}}
        			{\sqrt{\sum_{i} \sum_{t} (\bm{y}_{\bm{i}}^{\bm{t}} - \mathsf{mean}(\bar{\bm{y}}))^2}}
        \end{aligned}
        \end{equation}
        
    \item Empirical Correlation Coefficient (CORR):
        \begin{equation}
        \begin{aligned}
        CORR = \frac{1}{\bm{n}} \sum_{i=1}^{\bm{n}} \frac{\sum_{\bm{t}}  (\bm{y}_{\bm{i}}^{\bm{t}} - \mathsf{mean}({\bm{y}}_{\bm{i}}))
        (\bm{\bar{y}}_{\bm{i}}^{\bm{t}} - \mathsf{mean}(\bar{\bm{y}}_{\bm{i}}))}
        			{\sqrt{\sum_{i} \sum_{t} (\bm{y}_{\bm{i}}^{\bm{t}} - \mathsf{mean}(\bar{\bm{y}}))^2}}
        \end{aligned}
        \end{equation}
        
\end{itemize}


where $\bm{y}$ is the target variables values, $\bar{\bm{y}}$ is the predicted values, and $\bm{i}$ is the output feature index, and $\bm{t}$ is time stamp index of testing data. Throughout the analysis, we only present the testing errors.

\subsection{Kernel flow for time series forecasting}

In the this section, we  evaluate the performance of kernel flow model itself, introduced in Section 2.2, on time series data. We split data into 60\% for training, 20\% for validation, and 20\% for testing data. The results in Table \ref{table:1} compare the performance of kernel flow with the baseline models, based on the experimental results introduced in \cite{lai2018modeling}. The results show that the model can outperform most of the models in these time series benchmark data. We also examine this performance on power-load (PJM) data in Table \ref{tb:result-PJM}. The implemented LSTM-Attention model uses encoder-decoder architecture with Attention mechanism, similar to the model introduced in \cite{du2020multivariate}. A lower RSE and higher CORR are desired, where the best values are in bold font. The results on these two tables also show that kernel flow can be considered as a strong baseline model for time series forecasting problem.

\begin{table*}
\begin{center}
\caption{Comparison of kernel flow with baseline models on time series forecasting problem. Lower RSE and higher CORR are desired and presented in bold font.}
\scalebox{0.65}{
\begin{tabular}{ll|cccc|cccc|cccc|cccc}
\toprule
Dataset &  & \multicolumn{4}{c|}{Solar} & \multicolumn{4}{c|}{Traffic} & \multicolumn{4}{c|}{Electricity} & \multicolumn{4}{c}{Exchange-rate}  \\
\midrule
										 		&         & \multicolumn{4}{c|}{Horizon} & \multicolumn{4}{c|}{Horizon} & \multicolumn{4}{c|}{Horizon} & \multicolumn{4}{c}{Horizon} \\
\midrule
Methods 	& Metrics & 3 & 6 & 12 & 24 & 3 & 6 & 12 & 24 & 3 & 6 & 12 & 24 & 3 & 6 & 12 & 24 \\
\midrule
\multirow{1}{*}{LRidge} & RSE 
						              & 0.2019 & 0.2954 & 0.4832 & 0.7287 
                          & 0.5833 & 0.5920 & 0.6148 & 0.6025
                          & 0.1467 & 0.1419 & 0.2129 & 0.1280 
                          & 0.0184 & 0.0274 & 0.0419 & 0.0675\\
\textbf{\cite{lai2018modeling}}              & CORR 
                   		  & 0.9807 & 0.9568 & 0.8765 & 0.6803 
                          & 0.8038 & 0.8051 & 0.7879 & 0.7862
                          & 0.8890 & 0.8594 & 0.8003 & 0.8806 
                          & \textbf{0.9788} & \textbf{0.9722} & \textbf{0.9543} & \textbf{0.9305} \\   
\midrule
\multirow{1}{*}{GP} & RSE 
						  & 0.2259 & 0.3286 & 0.5200 & 0.7973
                          & 0.6082 & 0.6772 & 0.6406 & 0.5995
                          & 0.1500 & 0.1907 & 0.1621 & 0.1273 
                          & 0.0239 & 0.0272 & 0.0394 & 0.0580 \\
\textbf{\cite{lai2018modeling}}              & CORR 
                   		  & 0.9751 & 0.9448 & 0.8518 & 0.5971
                          & 0.7831 & 0.7406 & 0.7671 & 0.7909
                          & 0.8670 & 0.8334 & 0.8394 & 0.8818
                          & 0.8713 & 0.8193 & 0.8484 & 0.8278\\
\midrule
{\multirow{1}{*}{VARMLP}} & RSE 
						 	   & 0.1922 & 0.2679 & 0.4244 & 0.6841 
                               & 0.5582 & 0.6579 & 0.6023 & 0.6146 
                               & 0.1393 & 0.1620 & 0.1557 & 0.1274
                               & 0.0265 & 0.0304 & 0.0407 & 0.0578\\
\textbf{\cite{lai2018modeling}}                   & CORR 
                        	   & \textbf{0.9829} & 0.9655 & 0.9058 & 0.7149 
                               & 0.8245 & 0.7695 & 0.7929 & 0.7891
                               & 0.8708 & 0.8389 & 0.8192 & 0.8679
                               & 0.8609 & 0.8725 & 0.8280 & 0.7675 \\
\midrule 
\multirow{1}{*}{RNN-GRU}	& RSE 
							  & 0.1932 & 0.2628 & 0.4163 & 0.4852
							  & 0.5358 & \textbf{0.5522} & \textbf{0.5562} & \textbf{0.5633}
                			  & \textbf{0.1102} & \textbf{0.1144} & \textbf{0.1183} & \textbf{0.1295}
							  & 0.0192 & 0.0264 & 0.0408 & 0.0626 \\
\textbf{ \cite{lai2018modeling} }               & CORR 
                       		  & 0.9823 & 0.9675 & 0.9150 & 0.8823 
                              & \textbf{0.8511} & \textbf{0.8405} & \textbf{0.8345} & \textbf{0.8300}
                              & 0.8597 & 0.8623 & 0.8472 & 0.8651
                              & 0.9786 & 0.9712 & 0.9531 & 0.9223 \\

\midrule                                                
\multirow{1}{*}{KF}	& RSE 
							  & \textbf{0.1902} & \textbf{0.2517} & \textbf{0.3314} & \textbf{0.4537}
							  & \textbf{0.5290} & 0.5827 & 0.6148 & 0.6291
                			  & 0.1413 & 0.1445 & 0.1485 & 0.1542
							  & \textbf{0.0176} & \textbf{0.0243} & \textbf{0.0350} & \textbf{0.0470} \\
\textbf{Ours}               & CORR 
                       		  & 0.9827 & \textbf{0.9679} & \textbf{0.9419} & \textbf{0.8853} 
                              & 0.8304 & 0.7953 & 0.7702 & 0.7601
                              & \textbf{0.9111} & \textbf{0.9053} & \textbf{0.8977} & \textbf{0.8862}
                              & 0.9660 & 0.9355 & 0.8659 & 0.7598 \\

\bottomrule 
\end{tabular}}\label{table:1}
\end{center}
\end{table*}


                   
                   
                   
                   

                      
                   
                   

\begin{table}
\begin{center}
\caption{Comparison on PJM data}
\scalebox{0.9}{\begin{tabular}{ll|ccccc}
\toprule
										 		&         & \multicolumn{5}{c}{Horizon}  \\
\midrule
Methods 	& Metrics & 3 & 6 & 12 & 24 & 1-48 \\

\midrule
\multirow{1}{*}{LSTM-Attention}                 
           & MAPE 
                  & 4.02 & 5.65 & 7.16 & 7.62 & 7.93
                   \\ 
                   
\textbf{Ours}               
           & MAE 
                & 46.9 & 67.3 & 85.7 & 92.8 & 95.7  
                   \\
                   
\textbf{ }               
           & RMSE 
                  & 66.1 & 95.3 & 121.6 & 137.8 & 140.8
                   \\
       
\midrule
\multirow{1}{*}{Kernel-Flow}                 
           & MAPE 
                  & 3.94 & 5.88 & 7.18 & 7.68 & 7.95
                   \\ 
                   
\textbf{Ours}               
           & MAE 
                & 43.9 & 66.1 & 81.7 & 89.8 & 91.1 
                   \\
                   
\textbf{ }               
           & RMSE 
                  & 62.7 & 93.0 & 113.5 & 129.6 & 129.6
                   \\                   
\bottomrule

\end{tabular}}\label{tb:result-PJM}
\end{center}
\end{table}

\subsection{Over-fitting challenges on time series data} 

\begin{figure}[!ht]
\centering
    \includegraphics[width=2.0in]{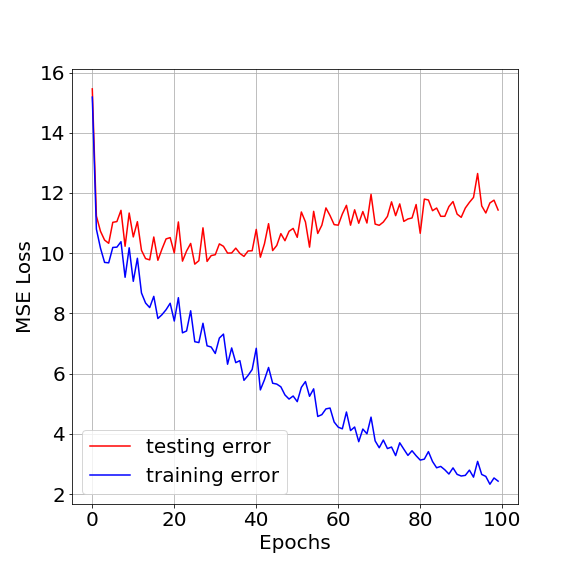}
\caption{Comparison of training and testing error at each epoch for a stacked LSTM without regularization.}
\label{fig:fig_train_test}
\end{figure}

In this section, we present the challenges related to overfitting of deep learning models on time series data. In Fig. \ref{fig:fig_train_test}, we illustrate the over-fitting of Stacked LSTM on power-load data. For both training and testing errors, RMSE, is obtained at the end of each epoch. The training error decreases throughout the training process and in all epochs. However, after 25 epochs, the testing error increases as the result of overfitting. We can also see that the minimum value of testing error is 9.63, while the testing error in the final epoch is 11.43, which nearly 18\% increases in testing error as the result of overfitting. We have two potential reasons for this big increase of testing error and overfitting. First, time series data have simpler patterns than images and text files. It is likely that an over-parametrized neural network would be overfitted to the training data. Second, LSTM layers are more vulnerable to overfitting than convolutional layers. A convolutional layer uses a weight sharing for the kernels. The kernel is generalized to local features in the input matrix, and it would not get overfitted easily. However, an LSTM layer has more vulnerability to overfitting, as it does not have a weight sharing. We examine this by adding various dropout rates to the LSTM layers, and we obtain the plot in Fig. \ref{fig:fig_droprates}. We can clearly see that dropout prevents overfitting to a good extent, and the model’s performance on testing data does not decrease. In the first 25 epochs, the testing error for both models with and without dropout are similar, but after epoch 25, the testing error for the model with dropout is lower than the model without dropout. Also, the best testing error with dropout is 9.24, and without dropout is 9.63, which 4.2\% improvement. The testing error after a fixed number of epochs, 100, are 11.41 and 10.00 for a model without dropout and with dropout, respectively, which is a 14.1\% improvement. 

\subsection{Hyper-parameter tuning} 

To evaluate regularization effects on LSTM neural networks, some hyper-parameters should be tuned to achieve better results. We have a grid-search for the following parameters. The look-back window (time window order) is selected from [12,24,36,48,72,96]. We found out on the hourly datasets, with daily seasonality, a look-back of 24 results in the best performance. Batch-size is selected as 256, and we observed that lower value of batch-size can de-stabilize training error. Before each epoch, training data is shuffled. Data shuffling can slightly change the results; hence, to have a reproducible result, we use fixed seeds for each epoch, similar in all the experiments. The weights of neural network are also randomly initialized with He-Normal initialization. We found out the weights of neural networks and data shuffling can slightly change our final results, as they results in randomness of the training process. Using seed numbers for both shuffling and model parameters, we can reproduce the results for all experiments. The number of LSTM layers is determined for each dataset using a grid-search on [1,2,3,4] layers. The final layer is a dense layer, which we found out one dense layer with a large number of units results in better performance.

\begin{figure}[!ht]
\centering
    \includegraphics[width=2.0in]{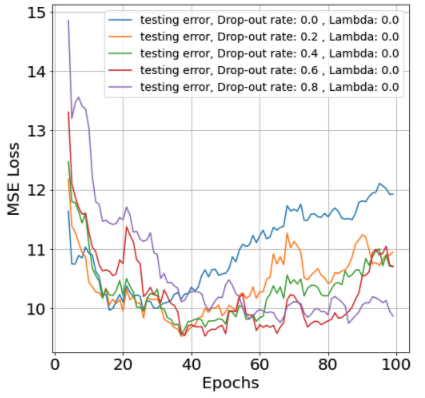}
\caption{Comparison of dropout-rates during training process.}
\label{fig:fig_droprates}
\end{figure}

The regularization parameters are Dropout-rate and Lambdas for kernel flow regularization. A grid search on dropout-rate of [0.0, 0.1, 0.2, 0.3, 0.4, 0.5, 0.6, 0.7, 0.8] is examined along with various number of units for LSTM layer [128, 256, 512], and the best parameters for both LSTM units and dropout-rate for each dataset are obtained. A subset of experiments on dropout-rates are presented in Fig. \ref{fig:fig_droprates}.

\begin{figure}[!ht]
\centering
    \includegraphics[width=2.0in]{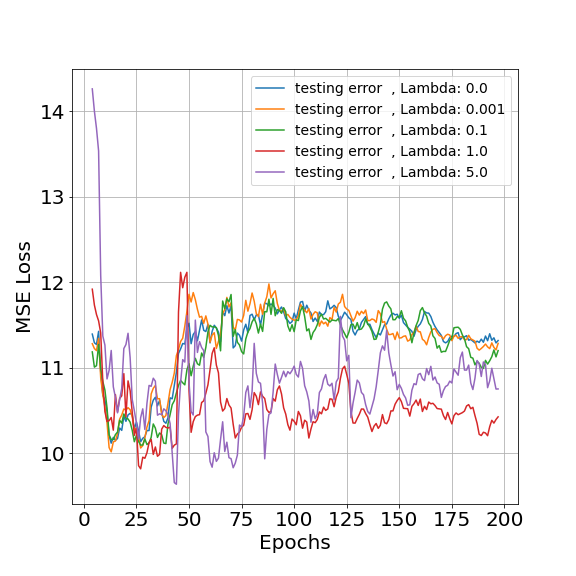}
\caption{Comparison of Lambdas for kernel flow regularization during training.}
\label{fig:fig_lambdas}
\end{figure}

\begin{figure*}[!ht]
\centering
    \subfloat[PJM power-load data\label{fig:output-b}]{\includegraphics[width=0.27\linewidth]{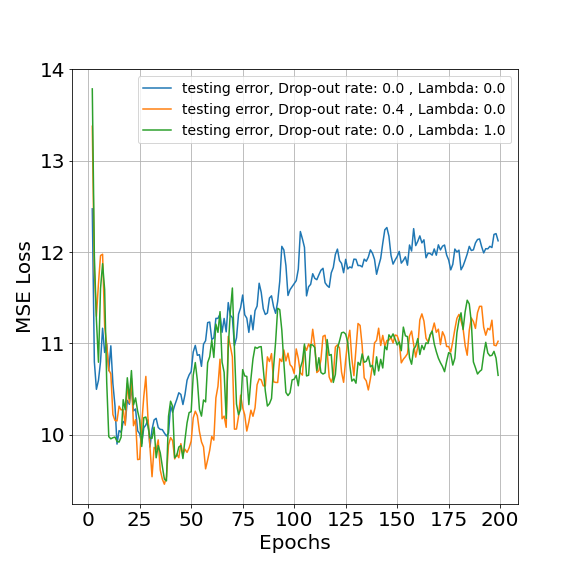}}\hfil
    \subfloat[Traffic data\label{fig:output-c}]{\includegraphics[width=0.27\linewidth]{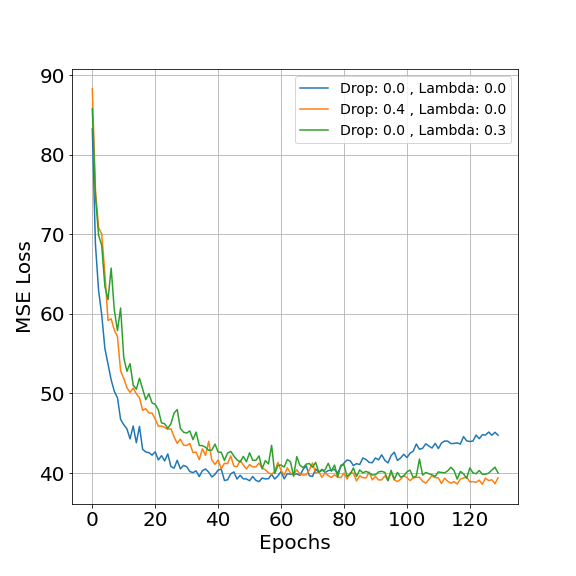}}\hfil
    \subfloat[Exchange-rate data\label{fig:output-d}]{\includegraphics[width=0.27\linewidth]{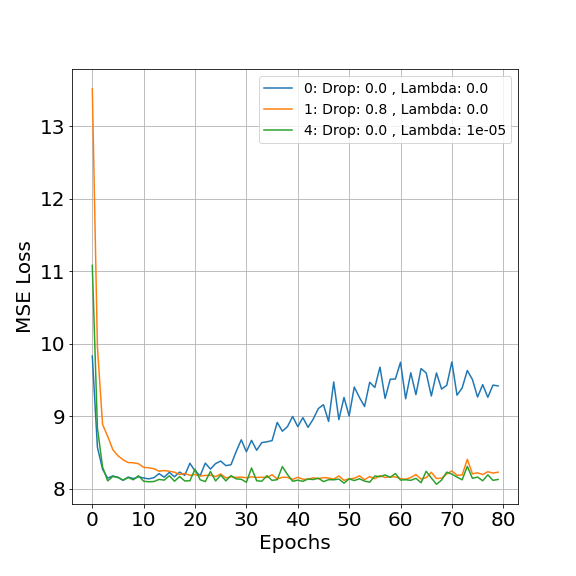}}
    
\caption{Comparison of testing error during training process.}
\label{fig:output-total}
\end{figure*}

Moreover, we examine Lambda values from the range of [0.001, 0.01, 0.1, 0.5, 1.0, 5.0, 10.0]. In Fig. \ref{fig:fig_lambdas}, the model without regularization, and the models with kernel flow regularization with lambdas 0.001 and 0.1 have similar testing error during the training process. It shows that a low value of Lambda does not have any effect, and the testing errors are similar with model without regularization. The model with lambda values of 1.0 has a significantly lower and stable error rate throughout 200 epochs of training. The lambda 5.0 shows the lowest error at epoch 40; however, it is not stabilized, and there are big fluctuations.

\begin{figure*}[!ht]

\centering
    \subfloat[No regularization]{\includegraphics[width=0.25\linewidth]{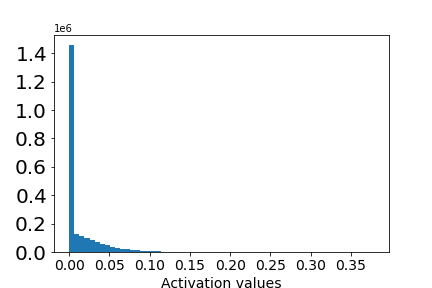}}\hfil
    \subfloat[With dropout]{\includegraphics[width=0.25\linewidth]{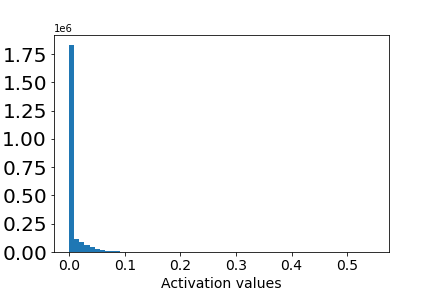}}
    \hfil
    \subfloat[with kernel flow regularization]{\includegraphics[width=0.25\linewidth]{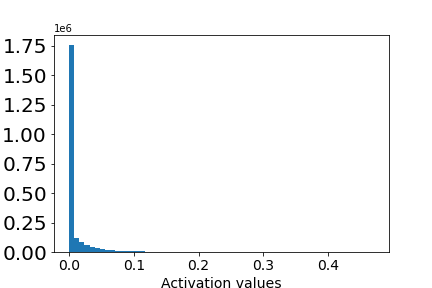}}
    
\caption{Histogram of feature vectors for regularization methods.}\label{fig:fig_higtogram_zeros}
\end{figure*}

\subsection{Comparison of regularization methods} 

In this section, we evaluate the effect of dropout and kernel flow regularization methods on training of an Stacked LSTM model. In Fig. \ref{fig:output-total}, the testing errors are presented for three datasets PJM, traffic and exchange-rate. Also, the results are for three models, Stacked LSTM without regularization, stacked LSTM with dropout, and stacked LSTM with kernel flow regularization. We can see that in all the three datasets, both dropout and kernel flow regularization have a comparable regularization effect on testing error. 
In Fig. \ref{fig:output-b}, the model with kernel flow regularization and dropout have the lowest testing error of 9.51 at epoch 38. While the model without regularization has the lowest testing error of 9.98 at epoch 19, which is nearly 5\% higher testing error than the other two models. Moreover, we can see that after 200 epochs, the model without regularization has nearly 10\% higher error than kernel flow regularization. In Fig. \ref{fig:output-c}, the analysis is applied to 50 traffic sensors. The model with kernel flow has the lowest testing error at epoch 116 with the value of 39.1, which is close to the best value of the model without regularization with the best testing error of 39.4. However, the model without regularization is overfitted after epoch 60. In Fig. \ref{fig:output-d}, the dataset is daily and significantly smaller than other datasets. The best errors of the models with/without regularization are almost similar. However, after epoch 25, the model without regularization is overfitted significantly, while both kernel flow regularization and dropout do not get overfitted even after 200 epochs. Their testing error is stabilized to the best error. The reasons can be related to the size of the dataset. In small datasets, regularization methods cannot find significantly better results. However, still there is big advantage of using kernel flow and dropout regularization methods. In real-world problems, the validation and testing data might be different. Hence, we cannot stop the training process at the lowest testing error, as we do not have the evaluation on testing data. As a result, in these scenarios, we need to have a fixed number of epochs, and stop the training process at the pre-defined step. In such a scenario, regularization methods have an important role to stabilize the testing error, and prevent overfitting.

Next, we compare the testing error of dropout and kernel flow regularizations for powerload forecasting problem. We use a grid-search method on dropout-rate and kernel flow weights, lambdas, to find the testing error obtained by early stopping method. We illustrate the models comparisons on power-load (PJM) data in Table \ref{grid-PJM}. For power-load data, we selected a network of 256 units for the LSTM, and 512 units for the fully-connected layer. We use dropout in the range of [0.0, 0.1, …, 0.7], and we use Lambdas from the range of [0.01, 0.05, 0.1, 0.5, 1.0, 5.0, 10.0]. There are totally 56 (8 by 7) trained model in this grid search. When Lambdas are greater than zero, then we have kernel flow regularization in the loss function. To better present the results in a brief table, we only present the best two models with both kernel flow and dropout, with only dropout and with only kernel flow. The results are in Table \ref{grid-PJM}. Totally, six models are in the table, which are represented with their rank among the trained models in grid search, their dropout-rate and Lambdas rate, and their RMSE. The results shows that the models with both dropout and kernel flow regularization methods achieves the lowest testing error during training process. The lowest error is for a model with both kernel flow and dropout regularization with RMSE of 40.78, which is better than the best model with only kernel flow regularization with RMSE of 41.51 and the best model with only dropout with RMSE of 41.94. In other words, the best model with kernel flow regularization has 2.8\% lower RMSE than the best model with dropout.

\subsection{Sparsity effect on features}

\begin{figure}[!ht]
\centering
    \includegraphics[width=2.2in]{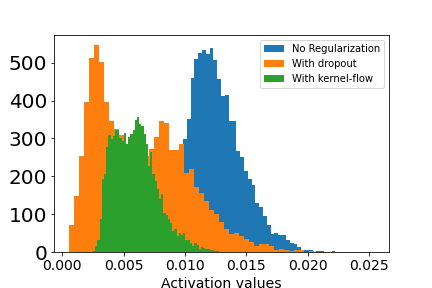}
\caption{Histogram of mean of feature vectors.}
\label{fig:fig_higtogram_mean}
\end{figure}

L1 and L2 norm regularization methods are effective in neural networks, as the activation of hidden units are sparse. In \cite{srivastava2014dropout}, it is shown that dropout automatically induces sparsity, and this property is mentioned as one of the effects of dropout regularization. We also evaluate the effect of kernel flow regularization on sparsity. We obtain the output of the last LSTM layer on testing data. Given $\bm{n}$ as the number of data points in testing data, and $\bm{h}$ as the hidden size, the output of LSTM layer is a vector of size $(\bm{n},\bm{h})$. Also, we consider a ReLU activation function at the last LSTM layer. Fig. \ref{fig:fig_higtogram_zeros} shows the histogram of activations for the models without regularization, with dropout, and with kernel flow regularization. We can see that the effect of dropout and kernel flow regularization are similar based on the sparsity of activations, that is the number of zeros as the output of ReLU activation is 1.75e+6 for both dropout and kernel flow, while it is 1.4e+6 without regularization. Moreover, we plotted the histogram of average of feature vectors for each data point. With a sparse vector, the average should be smaller \cite{srivastava2014dropout}. In Fig. \ref{fig:fig_higtogram_mean}, both kernel flow and dropout are results in lower average values, which is around 0.005 mean value, while the feature vectors of the model without regularization has the mean of 0.013. This analysis gives some insight of why kernel flow regularization is working. Additionally, the kernel flow sparsity is more concentrated around the mean activation value than drop-out method. This behaviour shows the sharpness of non-zero features, which is introduced as an important property in regularization of neural network layers in \cite{srivastava2014dropout}.


\vspace{0.5in}
\begin{table}
\begin{center}
\caption{Comparison of testing errors obtained by early stopping method. The ranking is based on the 56 experiments in the grid search.}
\scalebox{1.0}{
\begin{tabular}{ |p{0.8cm}||p{1.5cm}|p{1.0cm}|p{1.0cm}|  }
 \multicolumn{4}{c}{} \\
 \hline
 Rank & Dropout-rate & Lambda & RMSE \\
 \hline
  1  & 0.1 & 0.1 & 40.78\\
  2 & 0.2 & 0.05 & 40.97\\
  8 & 0.0 & 0.05 & 41.51\\
  9 & 0.0 & 0.1 & 41.60\\
  10 & 0.6 & 0.0 & 41.94\\
  11 & 0.4 & 0.0 & 42.03\\
 \hline
\end{tabular}}\label{grid-PJM}
\end{center}
\end{table}

\section{Conclusions and Future Work}
We presented the application of kernel flow for various time series forecasting use cases. We showed that the performance of kernel flow is comparable or superior from some of the baseline machine learning methods on time series data. Moreover, we applied the kernel flow regularization along side with dropout to LSTM neural networks to resolve the potential overfitting of these networks. We showed that kernel flow regularization is as effective as dropout by examining it on three time series datasets. We also show that the combination of both methods results in lower test error. 

In our future work, we would extend the application of kernels in deep learning for various other learning problems. For instance, we observe that kernels have some inherent distance properties that can be applied for more advanced network architectures in deep learning layers such as causal convolutional, attention neural networks and transformers. This may have further motivations for improvements in areas like highly noisy time series forecasting data where the signal is significantly masked. 


\section{Acknowledgement}
We would like to express our appreciation to Beyond Limits Inc's leadership team for allowing us to conduct this research and also providing us with financial support and much encouragement during the course of study. We also appreciate useful feedback from professor Houman Owhadi from Caltech, and Shahram Farhadi from Beyond Limits.

\bibliographystyle{IEEEtran}
\bibliography{ref}

\end{document}